\documentclass[letterpaper]{article} 
\usepackage[preprint]{aaai2027}  
\usepackage[hyphens]{url}  
\usepackage{graphicx} 
\usepackage{natbib}  
\usepackage{caption} 
\usepackage{xcolor}
\usepackage{tikz}
\usetikzlibrary{positioning, arrows.meta, calc, fit, backgrounds, shapes.geometric, decorations.pathreplacing}

\usepackage{algorithm}
\usepackage{algorithmic}

\usepackage{amsmath}
\usepackage{amssymb}
\usepackage{mathtools}
\usepackage{amsthm}

\usepackage[capitalize,noabbrev]{cleveref}

\theoremstyle{plain}

\DeclareMathOperator*{\argmax}{arg\,max}
\DeclareMathOperator*{\argmin}{arg\,min}

\newcommand{\chimera}{ChiMERa-Bench}
\newcommand{\contact}{\textbf{ConTact}}

\usepackage{newfloat}
\usepackage{listings}
\DeclareCaptionStyle{ruled}{labelfont=normalfont,labelsep=colon,strut=off} 
\floatstyle{ruled}
\newfloat{listing}{tb}{lst}{}
\floatname{listing}{Listing}

\usepackage{booktabs}

\title{ConTact: Contact-First Antibody CDR Design via Explicit Interface Reasoning}

\author{
    Mansoor Ahmed\textsuperscript{\rm 1,2}\corresponding,
    Spencer VonBank\textsuperscript{\rm 3},
    Nadeem Taj\textsuperscript{\rm 4},
    Avais Jan\textsuperscript{\rm 1},
    Sujin Lee\textsuperscript{\rm 2},
    Naila Jan\textsuperscript{\rm 5},
    Murray Patterson\textsuperscript{\rm 1}\corresponding
}
\affiliations{
    \textsuperscript{\rm 1}Georgia State University, Atlanta, USA\\
    \textsuperscript{\rm 2}Georgia Institute of Technology, Atlanta, USA\\
    \textsuperscript{\rm 3}DePauw University, Indiana, USA\\
    \textsuperscript{\rm 4}University of Engineering and Technology, Lahore, Pakistan\\
    \textsuperscript{\rm 5}Accenture, Chesterbrook, USA\\
    mahmed76@student.gsu.edu, mpatterson30@gsu.edu
}

\begin{document}

\maketitle

\begin{abstract}
Computational antibody CDR design methods condition on antigen structure to generate binding loops. Yet, the existing architectures conflate two fundamentally distinct sub-problems: identifying which CDR positions will contact the antigen, and selecting amino acids at those positions. This forces models to learn contact reasoning implicitly through uniform message passing, diluting antigen signal across all positions equally. We introduce \contact{}, a contact-then-act architecture that explicitly decomposes CDR design into three cascaded stages: learning surface complementarity fingerprints, predicting CDR-antigen contacts, and injecting contact-gated antigen features into the prediction head. A distance-biased cross-attention module encodes geometric priors favoring spatial neighbors, a contact-weighted cross-entropy loss concentrates gradient signal on binding-critical positions, and a standoff-aware contact-recovery loss places the designed loop against its native epitope. On the \chimera{} dataset, \contact{} surpasses the baselines on the binding interface and structural metrics (backbone RMSD, fraction of native contacts, interface RMSD, DockQ, and epitope F1) on all three splits, improving native-contact recovery by 11 to 16\%, while matching the best sequence recovery. The source code is available at: \url{https://github.com/mansoor181/ConTact.git}
\end{abstract}

\section{Introduction}
\label{sec:intro}

Antibodies bind antigens through their complementarity-determining regions (CDRs), six hypervariable loops whose sequence and structure determine binding specificity~\citep{chothia1987canonical}. Computational CDR design methods condition on antigen structure to generate sequences and backbone conformations for these loops~\citep{luo2022diffab,kong2022mean,kong2023dymean,wu2025raad}. Yet a growing body of evidence shows that existing methods largely fail to leverage antigen information. The predictions remain nearly unchanged when the antigen is removed~\citep{li2025benchmarking}, and BLOSUM substitution matrices explain model outputs as well as learned likelihoods~\citep{uccar2025blosum,chinery2024simple}.

We argue that a fundamental cause is architectural, where the current methods conflate two distinct sub-problems into a single prediction head. The first sub-problem is \emph{where} the CDR will contact the antigen, i.e., which CDR positions form binding interactions. The second is \emph{what} amino acids to place at those positions, given the local chemistry of the binding partner. Equivariant GNNs such as MEAN~\citep{kong2022mean} and RAAD~\citep{wu2025raad} propagate antigen information through uniform message passing that treats all antigen residues equivalently. Diffusion-based methods like DiffAb~\citep{luo2022diffab} concatenate antibody and antigen residues into a flat graph with only a fragment-type embedding to distinguish them. Even dyMEAN~\citep{kong2023dymean}, which uses a shadow paratope mechanism and edge distance prediction for contact-aware graph construction, does not use predicted contacts to modulate sequence and structure prediction. In all cases, the model must simultaneously discover which positions are binding-relevant and what residues belong there, with a uniform cross-entropy loss that allocates equal learning capacity to every position.

The CDR-antigen interface is inherently sparse. A CDR-H3 of length 10--25 typically forms only 5--15 contacts with the antigen, and the amino acid identity at contact positions is directly constrained by the chemistry of the binding partner. For example, hydrophobic pockets select for complementary hydrophobic CDR residues, while charged patches favor oppositely charged side chains~\citep{chothia1987canonical,li2025benchmarking}. The non-contact positions are primarily constrained by backbone geometry and loop stability. Therefore, treating these two classes of positions equally wastes learning capacity on the less informative non-contact positions.

We propose \contact{}, an encoder-decoder architecture that decomposes CDR design into three explicit stages, addressing the \emph{where} before the \emph{what}. First, the model learns surface complementarity fingerprints that characterize the local binding environment at each CDR position, inspired by molecular surface fingerprints~\citep{gainza2020deciphering,gainza2023novo}. Second, it predicts which CDR positions will contact the antigen using a supervised contact predictor. Third, it selectively injects local antigen features into the CDR representation, gated by the predicted contact confidence, so that antigen information flows preferentially to binding-critical positions. A distance-biased cross-attention module provides geometric inductive bias by favoring spatial neighbors, and a contact-weighted cross-entropy loss concentrates gradient signal on positions the model identifies as contacts.

Our main contributions are:
\begin{enumerate}
    \item We identify the conflation of contact identification and sequence prediction as a structural limitation of existing CDR design architectures, and propose the \emph{contact-first} design paradigm that decomposes these sub-problems into an explicit three-stage cascade.
    \item We introduce a contact-gated injection mechanism that selectively routes antigen information to binding-relevant CDR positions, preventing noise from distant antigen residues at non-contact positions, and places the designed loop against its native epitope.
    \item We demonstrate on \chimera{} that \contact{} achieves significant improvements in structural and binding interface metrics, such as RMSD, fnat, interface RMSD, DockQ, and epitope F1 on all three splits.
\end{enumerate}

\section{Related Work}
\label{sec:related}

\paragraph{Equivariant GNN methods.}
MEAN~\citep{kong2022mean} introduced multi-channel equivariant attention with alternating intra-chain and inter-chain segment layers for CDR design. dyMEAN~\citep{kong2023dymean} extended this with a shadow paratope mechanism that predicts inter-chain edge distances for dynamic graph construction, making it the closest existing work to contact-aware CDR design. RAAD~\citep{wu2025raad} uses a multi-relational graph representation and a contrastive specificity objective for binding specificity. EvoStruct~\cite{ahmed2026evostruct} is another recent approach that combines the strengths of protein language models (PLMs) with GNNs via a structural adapter. \contact{} differs from all these approaches in that it uses predicted contacts to directly modulate the sequence, structure, and binding prediction through gated injection and position-specific loss weighting.

\paragraph{Diffusion and flow methods.}
DiffAb~\citep{luo2022diffab} models CDR generation as a joint diffusion process over coordinates, orientations, and amino acid types. AbFlowNet~\citep{abir2025abflownet} extends this with flow matching and trajectory balance loss. AbMEGD~\citep{chen2025AbMEGD} and RADAb~\citep{wang2024radab} add retrieval-augmented and multi-expert components. dyAb~\citep{tan2025dyab} applies flow matching with structure relaxation. FlowDesign~\citep{wu2025flowdesign} identifies that standard Gaussian priors are poorly suited for CDR generation and replaces them with data-driven prior distributions. All these methods treat antigen conditioning as a flat concatenation of antibody and antigen residues with fragment-type embeddings, applying uniform attention without distinguishing contact from non-contact positions. \contact{} addresses this as a complementary limitation of the conditioning mechanism itself.

\paragraph{Antigen conditioning failures.}
Multiple studies have documented that existing CDR design methods fail to effectively use antigen information~\cite{ahmed2026agforce}. \citet{li2025benchmarking} showed that predictions remain nearly unchanged when the antigen is removed. \citet{uccar2025blosum} demonstrated that BLOSUM substitution matrices explain model outputs as well as learned likelihoods. \citet{chinery2024simple} found that simple computational methods can outperform deep learning in generating diverse, binder-enriched antibody libraries. The contact-first decomposition in \contact{} directly targets this problem by providing an explicit, supervised pathway for antigen information to reach the sequence and structure prediction heads.

\paragraph{Multi-state design paradigms.}
The idea of predicting binding-relevant features before designing sequences has precedent in broader protein design. MaSIF-seed~\citep{gainza2023novo} predicts favorable interaction sites on molecular surfaces using learned surface fingerprints, then designs binders targeting those sites. RFdiffusion~\citep{watson2023rfdiffusion} generates protein backbones first, then designs sequences with ProteinMPNN~\citep{dauparas2022proteinmpnn}. \contact{} applies a similar strategy at the residue-contact level, stated as: \textit{predict which CDR positions will contact the antigen, then condition sequence and structure design on those predictions}. Unlike MaSIF-seed, which operates on molecular surfaces in a separate pipeline, \contact{} performs contact prediction and sequence-structure co-design end-to-end within a single differentiable architecture.

\section{Preliminaries}
\label{sec:preliminaries}

\subsection{Task Definition}
\label{sec:task}

We adopt the formulation from \chimera{}~\citep{ahmed2026chimerabench}. Given an antigen structure $A = \{(s_j, \mathbf{x}_j) \mid j \in V_A\}$, an epitope specification $E \subseteq V_A$, and an antibody framework $F = \{(s_i, \mathbf{x}_i) \mid i \in V_\text{FR}\}$, the task is to design CDR residues $R = \{(s_k, \mathbf{x}_k) \mid k \in V_\text{CDR}\}$ that maximize the conditional likelihood subject to epitope contact constraints:
\begin{equation}
    R^* = \argmax_{R} \; p_\theta\!\bigl(R \mid A, E, F\bigr), \quad
    \text{s.t.} \;\; \mathcal{C}(R, A) \neq \emptyset ,
    \label{eq:task}
\end{equation}
where each residue has amino acid type $s_k \in \{1, \ldots, 20\}$ and C$\alpha$ coordinate $\mathbf{x}_k \in \mathbb{R}^3$. We denote by $\mathcal{C}(R, A) = \{j \in V_A \mid \exists\, k \in V_\text{CDR}\!: \|\mathbf{x}_k - \mathbf{x}_j\| < d_c\}$ the set of antigen residues contacted within cutoff $d_c$. We target CDR-H3, the most variable of the six loops and the primary determinant of antigen specificity~\citep{chothia1987canonical}, and the loop on which all current methods perform worst.

\subsection{Graph Construction}
\label{sec:graph}

We represent the antibody-antigen complex as a heterogeneous graph $\mathcal{G} = (V, \mathcal{E})$. The node set $V = V_\text{HC} \cup V_\text{LC} \cup V_A \cup V_\text{glob} \cup V_\text{vn}$ contains residue nodes from the heavy chain ($V_\text{HC}$), light chain ($V_\text{LC}$), and antigen ($V_A$), three global delimiter tokens ($V_\text{glob} = \{\text{BOH}, \text{BOL}, \text{BOA}\}$), and $N_\text{vn}$ virtual nodes~\citep{sestak2024vn_egnn}. Each residue node $i$ carries amino acid type $s_i \in \{1, \ldots, 20\}$ and four backbone atom coordinates $\mathbf{X}_i = [\mathbf{x}_i^\text{N}, \mathbf{x}_i^{\text{C}\alpha}, \mathbf{x}_i^\text{C}, \mathbf{x}_i^\text{O}] \in \mathbb{R}^{4 \times 3}$.

The edge set $\mathcal{E}$ is partitioned into 10 typed subsets that capture different structural relationships between amino acids. Within each chain, we construct \emph{radial edges} connecting all pairs within a C$\alpha$ distance cutoff, \emph{sequential edges} linking residues separated by one or two positions in primary sequence, and \emph{KNN edges} connecting each residue to its nearest spatial neighbors. Across chains, we add \emph{inter-chain radial edges} and \emph{inter-chain KNN edges} that enable direct communication between antibody and antigen residues. Moreover, three \emph{global-to-chain edges} connect the delimiter tokens to their respective chains. Two \emph{virtual node edge types} connect each virtual node bidirectionally to all epitope and all CDR residues. This creates a two-hop shortcut between epitope and CDR, directly addressing the over-squashing problem~\citep{alon2021bottleneck} where information from distant epitope residues dilutes through many layers of sequential message passing.

Each edge $(i, j)$ carries a feature vector $\mathbf{e}_{ij}$ encoding edge type (one-hot), relative position in local coordinate frames, pairwise distance RBFs between backbone atom pairs, a quaternion encoding of relative backbone orientation, and local frame direction features. Virtual node edges use learnable feature vectors rather than geometric features.


\subsection{Contact Definition}
\label{sec:contact_def}

We define a CDR residue $k$ as contacting the antigen if its C$\alpha$ atom lies within 8~\AA{} of any antigen C$\alpha$ atom~\citep{xue2015computational,ovchinnikov2014fnat}:
\begin{equation}
    c_k = \mathbb{1}\!\left[\min_{j \in V_A} \|\mathbf{x}_k - \mathbf{x}_j\| < 8\text{~\AA}\right]
    \label{eq:contact_label}
\end{equation}
The binary labels $c_k \in \{0, 1\}$ serve as supervision for the contact prediction stage and as weights in the contact-weighted sequence loss. The concurrent literature has explored antigen-antibody interaction prediction as an isolated bipartite link prediction problem~\cite{ahmed2026epiformer,liu2024asep,ahmed2025wallepp}.

\section{Method}
\label{sec:method}

\begin{figure*}
    \centering
    \includegraphics[width=0.99\linewidth]{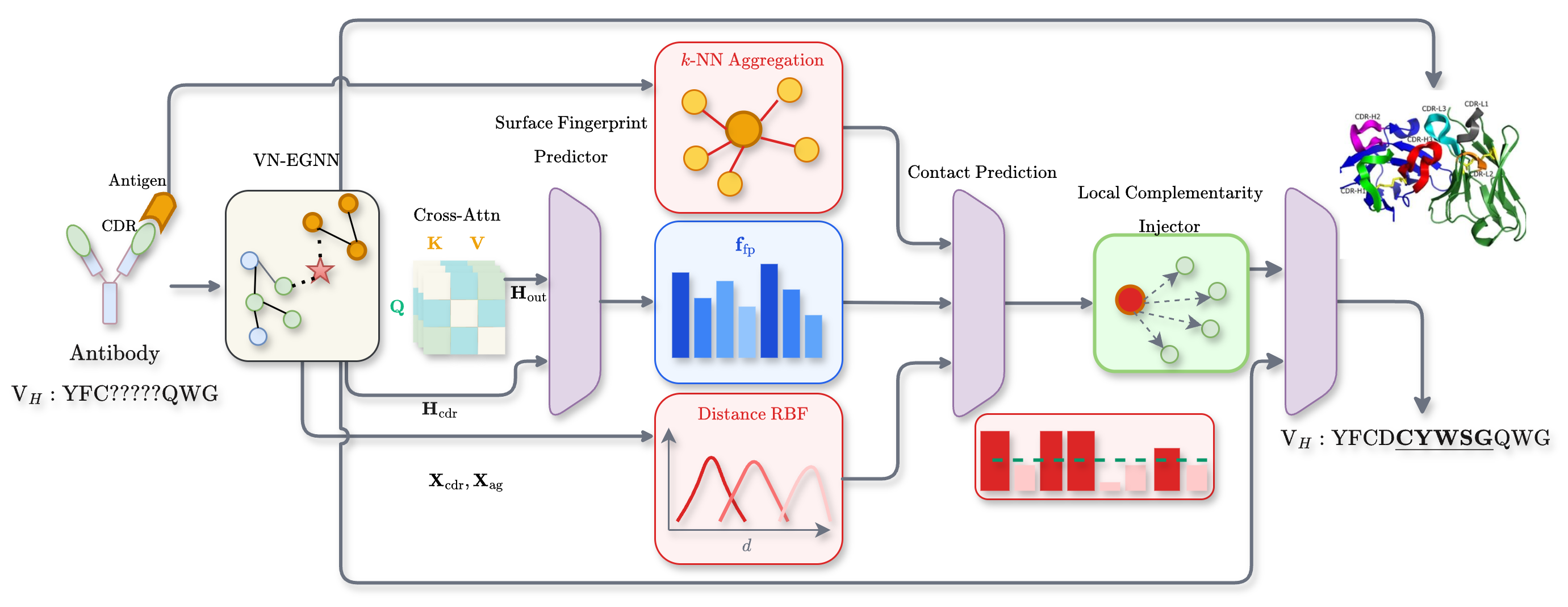}
    \caption{\textbf{\contact{} architecture.} The encoder maps residue features through a GNN to produce per-residue embeddings and updated coordinates. CDR and antigen embeddings are combined via distance-biased cross-attention. The three-stage decoder cascades complementarity fingerprinting, contact prediction, and contact-guided local complementarity injection. }
    \label{fig:architecture}
\end{figure*}

\contact{} consists of three components: (i) an EGNN encoder that performs E(3)-equivariant message passing over the heterogeneous graph, (ii) a distance-biased cross-attention module that combines CDR and antigen representations with spatial priors, and (iii) a three-stage contact-first decoder that cascades complementarity fingerprinting, contact prediction, and contact-guided sequence generation. \Cref{fig:architecture} illustrates the full architecture.

\subsection{Feature Encoding}
\label{sec:features}

Each residue $i$ in the antibody-antigen complex is represented by a feature vector $\mathbf{f}_i$ composed of five groups. \textbf{Amino acid identity}: a one-hot encoding of the residue type, masked to the zero vector for all CDR positions during training to prevent trivial teacher-forcing. \textbf{Backbone distance RBFs}: intraresidue bond lengths (N--C$\alpha$, C$\alpha$--C, C--O) each expanded into Gaussian basis functions:
\begin{equation}
    \phi_\text{rbf}(d)_m = \exp\!\left(-\frac{(d - \mu_m)^2}{2 \varsigma^2}\right), \quad m = 1, \ldots, M,
    \label{eq:rbf}
\end{equation}
where $\mu_m$ are uniformly spaced centers and $\varsigma$ is the basis width. \textbf{Backbone angles}: bond angles and dihedral angles ($\phi$, $\psi$, $\omega$), each encoded as sine-cosine pairs. \textbf{Local frame directions}: unit vectors along the three local coordinate axes defined by the N-C$\alpha$-C backbone triangle. \textbf{Sinusoidal position embedding}: encoding of the residue index within its chain at multiple frequency scales.


A segment type indicator distinguishes heavy chain, light chain, and antigen residues. A dual-path MLP processes geometric and chemical features through separate pathways with SiLU activations, fuses the outputs, and projects to embedding dimension $d$:
\begin{equation}
    \mathbf{h}_i^{(0)} = \text{MLP}_\text{fuse}\!\left([\text{MLP}_\text{geom}(\mathbf{f}_i^\text{geom}),\; \text{MLP}_\text{chem}(\mathbf{f}_i^\text{chem})]\right)
    \label{eq:feat_enc}
\end{equation}

Epitope residues (those in $E$) receive an additional learnable embedding $\mathbf{e}_\text{epi}$ added to their representation, providing an explicit signal that these residues are part of the designated binding site.

\subsection{EGNN Encoder}
\label{sec:encoder}

The encoder applies multiple relation-aware E(3)-equivariant GNN layers~\citep{satorras2021n} on graph $\mathcal{G}$. The virtual nodes (VN) with learnable feature vectors and learnable coordinates participate in message passing through the VN-to-epitope and VN-to-CDR edge types. Because each virtual node connects to all epitope and CDR residues, information flows from any epitope residue to any CDR residue in exactly two message-passing steps. Without virtual nodes, this information must traverse the graph via sequential edges, suffering from over-squashing~\citep{alon2021bottleneck} at bottleneck residues.

Each layer $l$ updates node features and coordinates simultaneously. For edge $(i, j)$ of type $t$, the message function takes the concatenation of sender and receiver embeddings, an outer product geometry term, and the edge features:
\begin{equation}
    \mathbf{m}_{ij}^{(l)} = \text{MLP}_\text{msg}^{(l)}\!\left([\mathbf{h}_i^{(l)},\; \mathbf{h}_j^{(l)},\; \text{vec}\!\left(\Delta\mathbf{x}_{ij} (\Delta\mathbf{x}_{ij})^\top\right),\; \mathbf{e}_{ij}]\right),
    \label{eq:message}
\end{equation}
where $\Delta\mathbf{x}_{ij} = \mathbf{x}_i^{(l)} - \mathbf{x}_j^{(l)}$ and $\text{vec}(\cdot)$ flattens the $3 \times 3$ outer product matrix into a 9-dimensional vector. The entries of $\Delta\mathbf{x}_{ij} (\Delta\mathbf{x}_{ij})^\top$ are dot products of displacement components, which are invariant to rotations, translations, and reflections.


The model aggregates messages from all edge types with type-specific linear projections and updates node features via a residual connection:
\begin{equation}
    \mathbf{h}_i^{(l+1)} = \mathbf{h}_i^{(l)} + \text{MLP}_\text{node}^{(l)}\!\left(\left[\mathbf{h}_i^{(l)},\; \sum_{t} \mathbf{W}_t^{(l)} \sum_{j \in \mathcal{N}_t(i)} \mathbf{m}_{ij}^{(l)}\right]\right),
    \label{eq:node_update}
\end{equation}
where $\mathbf{W}_t^{(l)}$ is a type-specific projection matrix and $\mathcal{N}_t(i)$ denotes the neighbors of node $i$ under edge type $t$. Coordinates are updated equivariantly by adding a weighted sum of displacement vectors:
\begin{equation}
    \mathbf{x}_i^{(l+1)} = \mathbf{x}_i^{(l)} + \sum_{t} \frac{1}{|\mathcal{N}_t(i)|}\sum_{j \in \mathcal{N}_t(i)} \Delta\mathbf{x}_{ij} \cdot \text{MLP}_t^{\text{coord},(l)}(\mathbf{m}_{ij}^{(l)}),
    \label{eq:coord_update}
\end{equation}
where $\text{MLP}_t^{\text{coord},(l)}$ produces a scalar weight. The product of a displacement vector and a scalar computed from invariant inputs is equivariant by construction. After all layers, the encoder produces residue embeddings $\mathbf{h} \in \mathbb{R}^{N \times D}$ and updated backbone coordinates $\hat{\mathbf{X}}$.

\subsection{Distance-Biased Cross-Attention}
\label{sec:cross_attn}

After encoding, we extract CDR node embeddings $\mathbf{H}_\text{cdr} \in \mathbb{R}^{L \times D}$ and antigen node embeddings $\mathbf{H}_\text{ag} \in \mathbb{R}^{M \times D}$. We note that standard cross-attention computes alignment scores purely from learned feature similarity, making no distinction between an antigen residue 5~\AA{} from the CDR and one 30~\AA{} away. \contact{} adds a Gaussian spatial bias that encodes a geometric inductive bias, since binding contacts are necessarily spatial neighbors.

We project queries $\mathbf{Q} = \mathbf{H}_\text{cdr}\mathbf{W}_Q$ and keys $\mathbf{K} = \mathbf{H}_\text{ag}\mathbf{W}_K$ into $H$ attention heads. The attention score between CDR position $i$ and antigen position $j$ is:
\begin{equation}
    \alpha_{ij}^{(h)} = \text{softmax}_j\!\left(\frac{(\mathbf{q}_i^{(h)})^\top \mathbf{k}_j^{(h)}}{\sqrt{D_h}} + \beta_{ij}\right),
    \label{eq:attn_score}
\end{equation}
where the distance bias $\beta_{ij}$ decays as a Gaussian function of the C$\alpha$--C$\alpha$ distance between the updated coordinates from the encoder:
\begin{equation}
    \beta_{ij} = \exp\!\left(-\frac{d_{ij}^2}{2\sigma^2}\right), \quad d_{ij} = \|\hat{\mathbf{x}}_i^{\text{C}\alpha} - \hat{\mathbf{x}}_j^{\text{C}\alpha}\|_2,
    \label{eq:dist_bias}
\end{equation}
with bandwidth $\sigma$. The Gaussian decay is near 1.0 for residues within van der Waals contact distance, drops to $e^{-2} \approx 0.14$ at $2\sigma$ (approximately the contact threshold), and becomes negligible beyond $3\sigma$. The bias is shared across heads to provide a consistent spatial prior, while the learned query-key projections specialize to different aspects of the binding interaction.

The output for each CDR position concatenates the multi-head weighted sum with the original CDR embedding:
\begin{equation}
    \mathbf{o}_i = \Big[\mathbf{h}_i^\text{cdr},\; \Big\|_{h=1}^{H} \sum_{j=1}^{M} \alpha_{ij}^{(h)} \mathbf{v}_j^{(h)}\Big],
    \label{eq:attn_out}
\end{equation}
where $\|$ denotes concatenation and $\mathbf{v}_j^{(h)} = \mathbf{h}_j^\text{ag} \mathbf{W}_V^{(h)}$ are value projections. The skip connection preserves the CDR embedding for downstream stages that primarily need structural context.

\subsection{Decoder}

\subsubsection*{Stage 1: Complementarity Fingerprinting}
\label{sec:fingerprint}

The first decoder stage compresses the cross-attention output into a compact representation of the local surface complementarity at each CDR position. The binding interactions follow chemical complementarity patterns (hydrophobic-hydrophobic, charge-charge, donor-acceptor) that can be captured in a low-dimensional fingerprint, analogous to molecular fingerprints in cheminformatics. Given the cross-attention output $\mathbf{o}_i$ for CDR position $i$, an MLP produces a fingerprint vector:
\begin{equation}
    \mathbf{f}_i = \text{MLP}_\text{fp}(\mathbf{o}_i)
    \label{eq:fingerprint}
\end{equation}

We train the fingerprint predictor with a contrastive loss that aligns CDR fingerprints with their local antigen environment. The antigen embeddings are projected to the fingerprint space via a separate MLP. For each contacting CDR position $i$, its nearest antigen neighbor serves as the positive, and distant antigen residues (beyond $1.5\times$ the contact cutoff) as negatives. The loss combines an InfoNCE~\citep{chen2020simple} term with a margin penalty on non-contact positions:

\begin{equation}
p_{ij^*} =
\frac{\exp(\mathbf{f}_i^\top \mathbf{g}_{j^*} / \tau_\text{fp})}
{\exp(\mathbf{f}_i^\top \mathbf{g}_{j^*} / \tau_\text{fp})
+ \sum_{k \in \mathcal{N}_i} \exp(\mathbf{f}_i^\top \mathbf{g}_k / \tau_\text{fp})},
\end{equation}

\begin{equation}
\mathcal{L}_\text{fp}
=
-\frac{1}{|\mathcal{C}|}\sum_{i \in \mathcal{C}} \log p_{ij^*}
+ \frac{1}{2}\mathcal{L}_\text{margin},
\label{eq:fp_loss}
\end{equation}


where $\mathcal{C}$ is the set of contacting CDR positions, $j^* = \argmin_j d_{ij}$ is the nearest antigen neighbor, $\mathbf{g}_j$ are the projected antigen fingerprints, $\mathcal{N}_i$ are the distant negatives, and $\mathcal{L}_\text{margin}$ penalizes non-contact positions whose maximum antigen similarity exceeds a margin threshold. The contrastive objective ensures that the fingerprint captures \emph{what kind of binding environment} a CDR position faces, conditioning the subsequent contact prediction stage.

\subsubsection*{Stage 2: Contact Prediction}
\label{sec:contact_pred}

The second stage predicts which CDR positions will form contacts with the antigen. This is the central component of the contact-first decomposition. Existing methods leave contact identification as an implicit byproduct of message passing~\citep{kong2022mean,wu2025raad} or graph construction~\citep{kong2023dymean}. \contact{} supervises contact prediction explicitly and uses the predictions to gate downstream information flow.

For each CDR position $i$, we aggregate features from its $K$ nearest antigen neighbors (by C$\alpha$ distance) using softmax-weighted attention over distances:
\begin{equation}
    \mathbf{a}_i = \sum_{j \in \text{KNN}_K(i)} w_{ij} \, \mathbf{h}_j^\text{ag}, \quad w_{ij} = \text{softmax}_j(-d_{ij}),
    \label{eq:knn_agg}
\end{equation}
where the softmax weights assign higher importance to closer neighbors. The contact predictor takes a concatenation of four inputs: the CDR embedding, the KNN-aggregated antigen features, an RBF encoding of the minimum distance to any antigen residue, and the complementarity fingerprint from Stage 1:
\begin{equation}
    \hat{c}_i = \sigma\!\left(\text{MLP}_\text{ct}\!\left([\mathbf{h}_i^\text{cdr},\; \mathbf{a}_i,\; \phi_\text{rbf}(d_i^\text{min}),\; \mathbf{f}_i]\right)\right),
    \label{eq:contact_pred}
\end{equation}
where $d_i^\text{min} = \min_{j \in V_A} \|\hat{\mathbf{x}}_i - \hat{\mathbf{x}}_j\|$ is the minimum C$\alpha$ distance to any antigen residue and $\sigma$ denotes the sigmoid function. Including the fingerprint $\mathbf{f}_i$ from Stage 1 creates a cascaded dependency, so that the quality of contact prediction depends on the learned complementarity representation.

The contact predictor outputs a soft probability $\hat{c}_i \in [0, 1]$ rather than hard binary predictions. The sigmoid output already provides smooth gradients, and hard thresholding introduced training instability in preliminary experiments. We supervise the contact predictor with a focal binary cross-entropy loss~\citep{lin2017focal} that addresses the inherent class imbalance, where non-contact positions typically outnumber contacts by 3--5$\times$:
\begin{equation}
    \mathcal{L}_\text{contact} = -\frac{1}{L}\sum_{i=1}^{L} (1 - \hat{p}_i)^\gamma \left[c_i \log \hat{c}_i + (1 - c_i) \log(1 - \hat{c}_i)\right],
    \label{eq:focal}
\end{equation}
where $c_i \in \{0, 1\}$ is the ground-truth contact label (\cref{eq:contact_label}), $\hat{p}_i$ denotes the predicted probability of the correct class, and $\gamma$ is the focusing parameter. The factor $(1 - \hat{p}_i)^\gamma$ down-weights well-classified examples, concentrating the learning signal on hard, ambiguous positions near the contact boundary.

\subsubsection*{Stage 3: Contact-Guided Local Complementarity Injection}
\label{sec:injection}

The third stage uses the predicted contact confidence $\hat{c}_i$ from Stage 2 to selectively inject local antigen information into the CDR embeddings. Antigen features should influence CDR representations primarily at positions the model predicts will form contacts, while non-contact positions should rely mainly on their backbone geometry context.

For each CDR position $i$, we aggregate features from $K$-nearest antigen neighbors using learned attention weights:
\begin{equation}
    \mathbf{h}_i^\text{local} = \sum_{j \in \text{KNN}_K(i)} \text{softmax}_j\!\left(\mathbf{w}_a^\top \mathbf{h}_j^\text{ag}\right) \mathbf{h}_j^\text{ag}
    \label{eq:local_agg}
\end{equation}
The contact confidence from Stage 2 directly gates the injection. The enriched CDR embedding combines the original representation with the contact-gated antigen information:
\begin{equation}
    \mathbf{h}_i^\text{enriched} = \mathbf{h}_i^\text{cdr} + \hat{c}_i \cdot \text{MLP}_\text{proj}(\mathbf{h}_i^\text{local})
    \label{eq:enriched}
\end{equation}
At non-contact positions ($\hat{c}_i \approx 0$), the antigen contribution is effectively zeroed out, preventing noise from distant antigen residues. At contact positions ($\hat{c}_i \approx 1$), the full projected local complementarity is injected. This single-gate design uses the contact predictor's output directly as the information bottleneck, avoiding the redundancy of additional learned gates.

The final representation for the sequence head concatenates the enriched embedding with a soft-masked cross-attention output:
\begin{equation}
    \mathbf{z}_i = [\mathbf{h}_i^\text{enriched},\; m_i \cdot \mathbf{o}_i^\text{attn}], \quad m_i = \epsilon + (1 - \epsilon)\,\hat{c}_i \quad,
    \label{eq:final_repr}
\end{equation}
where $\mathbf{o}_i^\text{attn}$ is the attention output from \cref{sec:cross_attn} and $\epsilon = 0.15$ is a minimum weight floor. Rather than fully zeroing attention at non-contact positions, the floor preserves a residual antigen signal that allows the model to recover from imperfect contact predictions.

\begin{table*}[h!]
\centering
\caption{CDR-H3 design performance comparison of \contact{} with the baseline methods on \chimera{} (antigen-fold split). Best in \textbf{bold}, second-best \underline{underlined}. Perplexity (PPL) is unavailable for diffusion and flow models.}
\label{tab:main_fold}
\small
\setlength{\tabcolsep}{4pt}
\begin{tabular}{lccccccc}
\toprule
Method & AAR$\uparrow$ & PPL$\downarrow$ & RMSD$\downarrow$ & fnat$\uparrow$ & iRMSD$\downarrow$ & DockQ$\uparrow$ & epiF1$\uparrow$ \\
\midrule
MEAN~\citep{kong2022mean} & \textbf{0.40} & \textbf{2.79} & 1.66 & \underline{0.62} & \underline{1.42} & \underline{0.70} & \underline{0.78} \\
dyMEAN~\citep{kong2023dymean} & 0.37 & 3.35 & 2.06 & 0.61 & 1.79 & 0.66 & 0.75 \\
RAAD~\citep{wu2025raad} & 0.38 & 3.06 & \underline{1.65} & \underline{0.62} & \underline{1.42} & \underline{0.70} & \underline{0.78} \\
dyAb~\citep{tan2025dyab} & 0.28 & -- & 2.40 & 0.56 & 1.81 & 0.65 & 0.72 \\
DiffAb~\citep{luo2022diffab} & 0.23 & -- & 2.24 & 0.52 & 2.10 & 0.58 & 0.65 \\
RADAb~\citep{wang2024radab} & 0.25 & -- & 7.79 & 0.51 & 4.78 & 0.57 & 0.65 \\
AbFlowNet~\citep{abir2025abflownet} & 0.23 & -- & 2.70 & 0.56 & 2.46 & 0.60 & 0.65 \\
AbMEGD~\citep{chen2025AbMEGD} & 0.24 & -- & 2.25 & 0.55 & 2.13 & 0.59 & 0.65 \\
AbODE~\citep{verma2023abode} & 0.27 & 10.78 & 13.17 & 0.16 & 4.00 & 0.39 & 0.40 \\
AbDockGen~\citep{jin2022hern} & 0.24 & 7.89 & 3.65 & 0.46 & 2.41 & 0.57 & 0.75 \\
\midrule
\textbf{\contact{}} (ours) & \underline{0.39} & \underline{3.01} & \textbf{1.59} & \textbf{0.68} & \textbf{1.32} & \textbf{0.73} & \textbf{0.81} \\
\bottomrule
\end{tabular}
\end{table*}

\subsubsection*{Contact-Weighted Sequence Head}
\label{sec:seq_head}

The sequence head maps the final representation $\mathbf{z}_i$ to amino acid logits $\boldsymbol{\ell}_i \in \mathbb{R}^{|\mathcal{V}|}$ via an MLP, where $\mathcal{V}$ includes the 20 standard amino acids plus special tokens (masked during inference). We apply label smoothing ($\epsilon_\text{ls} = 0.1$) and a contact-weighted variant of cross-entropy that allocates more learning capacity to positions predicted to form binding contacts:
\begin{equation}
    \mathcal{L}_\text{seq} = -\frac{1}{L}\sum_{i=1}^{L} w_i \, \text{CE}_{\epsilon_\text{ls}}(\boldsymbol{\ell}_i, y_i),
    \label{eq:weighted_ce}
\end{equation}
where $y_i$ is the ground-truth amino acid at position $i$ and the position-specific weight is:
\begin{equation}
    w_i = 1 + \alpha \cdot \hat{c}_i
    \label{eq:ce_weight}
\end{equation}
The hyperparameter $\alpha$ controls the relative up-weighting of contact positions. This reweighting follows from the observation that standard cross-entropy distributes learning capacity uniformly, treating a non-contact glycine at the loop apex the same as a contact-forming tryptophan buried in an antigen pocket. By up-weighting contacts, the model receives stronger gradient signal at precisely the positions where amino acid identity is most constrained by the antigen.

At inference, the predicted amino acid at each position is $\hat{s}_i = \argmax_a \ell_i^a$. The contact predictions $\hat{c}_i$ can also be inspected to verify which positions the model believes form contacts.

\subsection{Training Objective}
\label{sec:training}

The full training objective combines five active loss terms:
\begin{equation}
\begin{split}
    \mathcal{L} = \mathcal{L}_\text{seq} &+ \lambda_\text{coord}\mathcal{L}_\text{coord} + \lambda_\text{contact}\mathcal{L}_\text{contact} \\
    &+ \lambda_\text{fp}\mathcal{L}_\text{fp} + \lambda_\text{crl}\mathcal{L}_\text{crl}
\end{split}
    \label{eq:full_loss}
\end{equation}
Two terms are load-bearing infrastructure (the sequence cross-entropy of \cref{eq:weighted_ce} and the coordinate loss), and the remaining three carry the contact contribution. The coordinate loss $\mathcal{L}_\text{coord}$ is a smooth-$\ell_1$ (Huber) loss on predicted versus true C$\alpha$ coordinates for CDR positions:
\begin{equation}
    \mathcal{L}_\text{coord} = \frac{1}{L}\sum_{k \in V_\text{CDR}} \text{smooth}_{\ell_1}\!\left(\hat{\mathbf{x}}_k^{\text{C}\alpha} - \mathbf{x}_k^{\text{C}\alpha,\text{true}}\right)
\end{equation}

The contact-recovery loss $\mathcal{L}_\text{crl}$ supervises where the designed loop sits relative to the antigen, rather than where each residue sits in absolute space. For each CDR position $k$ that contacts the antigen in the native complex, let $d_k = \min_{j \in \mathcal{C}_k} \|\hat{\mathbf{x}}_k - \mathbf{x}_j\|$ be the distance from the predicted C$\alpha$ to the nearest of its native epitope partners $\mathcal{C}_k$. A one-sided hinge pulls this distance under the contact cutoff, and a clash floor prevents the loop from collapsing into the antigen:
\begin{equation}
    \mathcal{L}_\text{crl} = \frac{1}{|\mathcal{C}|}\sum_{k \in \mathcal{C}} \Big[ \max(0,\, d_k - d_c + \delta) + \max(0,\, d_\text{min} - d_k) \Big],
    \label{eq:crl}
\end{equation}
where $d_c = 8$~\AA{} is the contact cutoff, $\delta = 0.5$~\AA{} is a standoff margin that requires the prediction to clear the boundary rather than sit on it, and $d_\text{min} = 4$~\AA{} is the clash floor. Because the objective constrains a relative distance to the binding partner instead of an absolute coordinate, it improves interface placement without over-constraining the loop conformation.
All loss weights $\lambda$ are determined by hyperparameter sweeps using Weights \& Biases (W\&B).

\section{Experiments}
\label{sec:experiments}

\subsection{Setup}

\paragraph{Dataset and metrics.} We evaluate on \chimera{}~\citep{ahmed2026chimerabench}, comprising 2,922 antibody-antigen complexes, across three pre-defined splits. The antigen-fold split holds out antigen structural folds, the epitope-group split clusters complexes by epitope so that test epitopes are unseen during training, and the temporal split trains on older structures and tests on newer ones. Each split uses roughly 2,338 training, 292 validation, and 292 test complexes. \contact{} designs CDR-H3, the most variable loop and the primary determinant of antigen specificity, so all reported results are for CDR-H3.

Sequence quality is measured by amino acid recovery (AAR) and perplexity (PPL). Structural quality is measured by C$\alpha$ RMSD. Binding quality is measured by fraction of native contacts (fnat), interface RMSD (iRMSD), DockQ~\citep{basu2016dockq}, and epitope F1. The implementation details are provided in Appendix~\ref{app:imp}.

\paragraph{Baselines.} We compare against eight baselines spanning different architectural families: equivariant GNNs (RAAD~\citep{wu2025raad}, MEAN~\citep{kong2022mean}, dyMEAN~\citep{kong2023dymean}), diffusion and flow models (DiffAb~\citep{luo2022diffab}, AbFlowNet~\citep{abir2025abflownet}, AbMEGD~\citep{chen2025AbMEGD}, dyAb~\citep{tan2025dyab}), and autoregressive models (AbDockGen~\citep{jin2022hern}). All models are retrained on the \chimera{} dataset.

\subsection{Main Results }

\Cref{tab:main_epi,tab:main_fold,tab:main_temporal} present the performance comparison of \contact{} with the baselines across the three splits. \contact{} attains the best value on every structural and binding metric on all three splits. 

\paragraph{Structural quality.} \contact{} produces the most accurate backbones on CDR-H3, the hardest and most variable loop. It achieves the lowest RMSD on every split (1.58~\AA{} on antigen-fold, 1.72~\AA{} on temporal, 1.90~\AA{} on epitope-group), improving over the strongest baseline by 4.2\% on antigen-fold (1.58 vs RAAD 1.65~\AA), 5.0\% on temporal (1.72 vs RAAD 1.81~\AA), and 2.6\% on epitope-group (1.90 vs RAAD 1.95~\AA). The margin over the diffusion and flow models is considerably larger, exceeding 25\% on every split. Coordinate supervision over the equivariant encoder anchors the backbone in absolute space, and the contact-recovery loss then places it relative to the epitope, which is what separates \contact{} from baselines that optimize coordinates alone.

\paragraph{Binding interface quality.} The interface metrics separate \contact{} from the baselines more sharply than RMSD does. \contact{} achieves the best fnat on all three splits (0.58 on epitope-group, 0.69 on antigen-fold, 0.69 on temporal), a relative improvement of 16.0\%, 11.3\%, and 13.1\% over the strongest baseline on each split. It also achieves the best iRMSD (1.60, 1.31, and 1.38~\AA) and the best DockQ (0.67, 0.73, and 0.74, each three points above the best baseline). The gap is widest on fnat, the metric that measures how many native contacts the designed loop actually reproduces, which is precisely the quantity the contact cascade and the contact-recovery loss optimize. Baselines that route antigen information through uniform message passing (MEAN, RAAD) reach comparable backbone accuracy but recover 7 to 10 fewer native contacts per 100, so their loops are correctly shaped yet imprecisely placed against the epitope.

\paragraph{Epitope awareness.} \contact{} achieves the best epitope F1 on all three splits (0.69, 0.82, and 0.78), surpassing the next-best baseline by 2 points on epitope-group and 4 points on both antigen-fold and temporal. Epitope F1 measures whether the designed loop contacts the intended antigen residues rather than an adjacent surface patch, so it isolates antigen conditioning from loop geometry. The ablations attribute this gain to the contact cascade rather than to the antigen branch alone.

\paragraph{Sequence recovery.} \contact{} achieves an AAR of 0.40 to 0.41, the best on antigen-fold and temporal and within one point of MEAN on epitope-group. Sequence recovery on CDR-H3 is therefore bounded by a strong positional prior that all current methods operate close to, while the interface and structural metrics remain the axis on which architectures separate. We additionally conduct thorough diagnostic experiments to test the sequence recovery on this task and found bottlenecks for the GNN methods (full details in Appendix~\ref{sec:placement_wall}).

\subsection{Ablation Study}
\label{sec:ablation}

\Cref{tab:ablation} isolates each design choice on the epitope-group split. The reference row is the contact-recovery loss alone with the cascade disabled. We use this configuration as a reference and remove the components from \contact{} to compare their contribution.

\begin{table}[h!]
\centering
\caption{Ablations on \chimera{} (CDR-H3, epitope-group split). Each row changes one design choice relative to the contact-recovery-only reference. Best in \textbf{bold}.}
\label{tab:ablation}
\scriptsize
\setlength{\tabcolsep}{1.8pt}
\begin{tabular}{lccccccccc}
\toprule
Config & Casc. & $\alpha$ & Germ. & AAR & RMSD & fnat & iRMSD & DockQ & epiF1 \\
\midrule
Reference (CRL only) & $\times$ & 0 & \checkmark & 0.407 & \textbf{1.887} & 0.565 & \textbf{1.537} & 0.667 & 0.705 \\
\textbf{$+$ cascade (full)} & \checkmark & 3 & \checkmark & 0.407 & 1.912 & \textbf{0.593} & 1.586 & \textbf{0.674} & \textbf{0.724} \\
\midrule
Cascade, $\alpha=2$ & \checkmark & 2 & \checkmark & \textbf{0.408} & 1.930 & 0.539 & 1.617 & 0.653 & 0.681 \\
Interface distogram & $\times$ & 0 & \checkmark & 0.407 & 1.976 & 0.579 & 1.651 & 0.662 & 0.700 \\
Cascade, no germline & \checkmark & 3 & $\times$ & 0.377 & 1.909 & 0.579 & 1.573 & 0.670 & 0.719 \\
\quad $+$ KL to repertoire & \checkmark & 3 & $\times$ & 0.367 & 1.888 & 0.585 & 1.587 & 0.670 & 0.706 \\
\bottomrule
\end{tabular}
\end{table}

\paragraph{Cascade improves the interface metrics.} 
Enabling the three-stage cascade, together with its contact and fingerprint auxiliary losses, raises fnat from 0.565 to 0.593, epitope F1 from 0.705 to 0.724, and DockQ from 0.667 to 0.674, at a flat AAR and a 0.05~\AA{} cost in iRMSD. The gain comes from the auxiliary losses shaping the shared encoder rather than from the contact-weighted cross-entropy, which is inert when the cascade is off: a disabled contact head emits a constant confidence of 0.5, so every position receives the same weight. This is the central ablation result. The contact-first decomposition buys interface quality, and it does not claim to buy sequence recovery.

\paragraph{Higher contact weight.} Lowering $\alpha$ from 3 to 2 regresses every interface metric (fnat 0.539, iRMSD 1.617~\AA, DockQ 0.653, epitope F1 0.681) while leaving AAR unchanged. A weak reweighting spreads gradient back toward non-contact positions and reduces the benefit of the cascade.

\paragraph{An exact-distance objective does not substitute for the cascade.} Replacing the cascade with an interface distogram loss, which matches predicted CDR-epitope C$\alpha$--C$\alpha$ distances to their native values, holds fnat and DockQ at the reference level while worsening both RMSD (1.976~\AA) and iRMSD (1.651~\AA), the two metrics it most directly supervises. Matching exact interface distances is a harder objective than the placement hinge and does not escape the same contour.


\paragraph{Limitations.} \contact{} improves where the designed loop sits against the antigen, but it does not translate that improvement into antigen-specific residue identities. The contact predictor is also imperfect, and the model conditions on C$\alpha$ geometry alone. Future work should couple a sharper apex generator, or an external evolutionary prior, back into the sequence head so that improved placement can inform residue choice.

\section{Conclusion}
\label{sec:conclusion}

In this paper, we proposed \contact{}, which decomposes antibody CDR design into contact prediction followed by contact-guided sequence-structure co-generation. The three-stage cascade of complementarity fingerprinting, contact prediction, and contact-gated injection provides a supervised pathway for antigen information to reach the prediction head at binding-relevant positions. Extensive experiments on \chimera{} across three splits demonstrate that this contact-first decomposition attains the best structural and binding interface metrics, while matching the best sequence recovery.






\newpage

\bibliography{references}

\onecolumn
\newpage
\appendix
\section{Appendix}

\subsection{Implementation details.} \label{app:imp}

This section presents the training configuration for \contact{}. Specifically, the feature encoder projects 108D input features to 32D embeddings. The EGNN encoder uses 4 layers with 256D hidden features and 3 virtual nodes. The cross-attention module uses 2 heads with bandwidth $\sigma = 4$~\AA. The complementarity fingerprint is 32-dimensional, while the contact predictor MLP has hidden layers of 256 and 128 units with LayerNorm. The loss weights are $\lambda_\text{coord} = 0.3$, $\lambda_\text{contact} = 1.0$, $\lambda_\text{fp} = 0.1$, and $\lambda_\text{crl} = 1.0$, determined by Weights \& Biases sweep. Furthermore, the contact weight $\alpha = 3$ and the focal parameter $\gamma = 2$. We train with Adam (lr $= 6.31 \times 10^{-4}$, exponential decay $\gamma_\text{lr} = 0.944$ per epoch), gradient clipping at 0.5, batch size 8, dropout 0.1, and early stopping on validation loss with patience 10 for at most 100 epochs. A single run takes approximately 2.5 hours on one NVIDIA A100 GPU.
Moreover, all interface metrics use symmetric C$\alpha$--C$\alpha$ contacts at 8~\AA{} restricted to CDR residues.

\subsection{Additional results}
We further compare \contact{} with the baselines on the other two splits proposed in the \chimera{} dataset, namely temporal and epitope group splits. They are provided in Table~\ref{tab:main_temporal} and Table~\ref{tab:main_epi}, respectively.

\begin{table*}[h!]
\centering
\caption{CDR-H3 design on \chimera{}, temporal split ($n=293$). Best in \textbf{bold}, second-best \underline{underlined}.}
\label{tab:main_temporal}
\small
\setlength{\tabcolsep}{4pt}
\begin{tabular}{lccccccc}
\toprule
Method & AAR$\uparrow$ & PPL$\downarrow$ & RMSD$\downarrow$ & fnat$\uparrow$ & iRMSD$\downarrow$ & DockQ$\uparrow$ & epiF1$\uparrow$ \\
\midrule
MEAN & \underline{0.39} & 3.31 & 1.84 & 0.60 & 1.44 & 0.71 & 0.73 \\
RAAD & 0.38 & \underline{3.24} & 1.81 & 0.61 & 1.46 & 0.71 & 0.74 \\
dyMEAN & 0.35 & 3.28 & 2.32 & 0.50 & 2.00 & 0.63 & 0.64 \\
dyAb & 0.29 & -- & 2.89 & 0.51 & 1.92 & 0.64 & 0.65 \\
DiffAb & 0.23 & -- & 2.46 & 0.51 & 2.18 & 0.61 & 0.61 \\
AbFlowNet & 0.23 & -- & 2.67 & 0.52 & 2.33 & 0.61 & 0.61 \\
RADAb & 0.22 & -- & 11.57 & 0.48 & 7.91 & 0.57 & 0.57 \\
AbMEGD & 0.23 & -- & 2.72 & 0.55 & 2.35 & 0.62 & 0.58 \\
AbODE & 0.37 & 9.26 & 15.63 & 0.07 & 4.65 & 0.35 & 0.19 \\
AbDockGen & 0.23 & 7.55 & 3.89 & 0.37 & 2.52 & 0.55 & 0.68 \\
\midrule
\textbf{\contact{}} (ours) & \textbf{0.40} & \textbf{3.09} & \textbf{1.72} & \underline{0.66} & \textbf{1.38} & \underline{0.73} & \underline{0.77} \\
\bottomrule
\end{tabular}
\end{table*}

\begin{table*}[h!]
\centering
\caption{CDR-H3 design on \chimera{}, epitope-group split. Best in \textbf{bold}, second-best \underline{underlined}. }
\label{tab:main_epi}
\small
\setlength{\tabcolsep}{4pt}
\begin{tabular}{lccccccc}
\toprule
Method & AAR$\uparrow$ & PPL$\downarrow$ & RMSD$\downarrow$ & fnat$\uparrow$ & iRMSD$\downarrow$ & DockQ$\uparrow$ & epiF1$\uparrow$ \\
\midrule
MEAN & \textbf{0.42} & \textbf{3.00} & 2.01 & 0.48 & 1.66 & 0.63 & 0.67 \\
RAAD & 0.38 & 3.31 & 1.95 & 0.49 & 1.64 & \underline{0.64} & 0.67 \\
dyAb & 0.27 & -- & 3.31 & 0.35 & 2.27 & 0.54 & 0.50 \\
DiffAb & 0.22 & -- & 2.64 & 0.48 & 2.27 & 0.58 & 0.56 \\
AbFlowNet & 0.22 & -- & 2.70 & 0.49 & 2.37 & 0.57 & 0.55 \\
RADAb & 0.22 & -- & 12.28 & 0.47 & 5.82 & 0.57 & 0.57 \\
AbMEGD & 0.22 & -- & 2.76 & \underline{0.50} & 2.41 & 0.57 & 0.56 \\
AbODE & 0.31 & 9.99 & 16.40 & 0.08 & 5.12 & 0.34 & 0.20 \\
AbDockGen & 0.25 & 7.41 & 3.97 & 0.35 & 2.72 & 0.52 & 0.63 \\
\midrule
\textbf{\contact{}} (ours) & \underline{0.41} & \underline{3.08} & \textbf{1.90} & \textbf{0.58} & \textbf{1.60} & \textbf{0.67} & \textbf{0.69} \\
\bottomrule
\end{tabular}
\end{table*}

\subsection{Why sequence recovery saturates (\textit{negative results})}
\label{sec:placement_wall}

Every GNN architecture in \cref{tab:main_epi,tab:main_fold,tab:main_temporal} achieves within a few points of the same sequence recovery. \contact{} improves the interface and structural metrics while performing competitively on AAR. Both the sequence ceiling and the interface ceiling trace to a single cause. Our experiments show that none of the models in this setting resolves the CDR-H3 apex backbone to better than about 2.4~\AA{} from the framework, the antigen, and the masked loop alone. We ran four independent probes on CDR-H3 and the epitope-group split to establish this.

\paragraph{The missing sequence signal is in the antibody's own backbone.} We fit an inverse-folding multilayer perceptron that predicts each H3 residue from geometric features, and we measure recovery separately at the apex (middle third of the loop), at native contacts, and at the J-anchor. \Cref{tab:probe} reports the result. Providing the \emph{ground-truth} H3 backbone lifts apex and contact recovery by 15 to 16 points over what position, length, and germline statistics provide, whereas a shuffle control that breaks the structure-sequence pairing collapses the gain to the position-only floor. The signal is therefore real geometry rather than the model capacity. We can draw the conclusion that the sequence is recoverable given an accurate apex backbone.

\begin{table}[h!]
\centering
\caption{Inverse-folding probe on CDR-H3 (epitope-group split). Recovery is reported overall and restricted to apex, native-contact, and J-anchor positions. The shuffle control breaks the structure-sequence pairing.}
\label{tab:probe}
\small
\setlength{\tabcolsep}{4pt}
\begin{tabular}{lcccc}
\toprule
Features & Overall & Apex & Contact & Anchor \\
\midrule
Position only & 0.328 & 0.177 & 0.186 & 0.519 \\
$+$ true torsions & 0.419 & 0.280 & 0.276 & 0.606 \\
$+$ local Cartesian & \textbf{0.461} & \textbf{0.339} & \textbf{0.333} & \textbf{0.629} \\
\midrule
Shuffle control & -- & 0.177 & 0.184 & -- \\
\bottomrule
\end{tabular}
\end{table}

\paragraph{The predicted backbone is worst specifically at the apex.} Breaking C$\alpha$ error down by loop decile shows that \contact{} builds accurate loop ends around an inaccurate middle. The apex averages 2.41~\AA{} while the last three J-anchor positions average 1.12~\AA. By cross-referencing the probe above, the structure-to-sequence signal turns net negative once backbone error exceeds 2~\AA. The model builds precisely the segment it cannot then read a sequence from, which is why the apex recovery stalls near 0.18.

\paragraph{Backbone co-design does not break the wall.} We implemented an inverse folding approach using the Natural Extension of Reference Frame (NeRF) algorithm~\cite{parsons2005nerf}, which converts internal coordinates (bond lengths, bond angles, and torsion/dihedral angles) into 3D Cartesian coordinates (x, y, z atomic positions). A torsion head predicts per-residue backbone dihedrals and bond angles, rebuilds the loop by NeRF anchored on the given framework, and feeds the resulting geometry embedding into the sequence head. The parametrization is exact, with a round-trip C$\alpha$-RMSD of 0.06~\AA{} under true internal coordinates, so any failure is predictive rather than representational. The head learns apex torsions only to 47$^\circ$ mean error, far short of the roughly 15$^\circ$ needed for sub-Angstrom placement, and feeding those torsions to the sequence head yields no gain (validation AAR 0.399, matching the no-torsion baseline). A local-only variant that drops the coordinate loss is equally flat, which rules out training instability.

\paragraph{The apex conformation is not retrievable.} If a near-native apex existed in the training library, a retrieval or multi-hypothesis scheme could place it, but it does not. For each test loop, we superimpose every same-length training H3 onto the test framework stem, using Kabsch alignment on five flanking C$\alpha$ per side, which is available at inference. An oracle that selects the best training loop reaches an apex error of 2.05~\AA{} median, essentially matching plain mean regression at 2.41~\AA. Hence, ranking by framework-stem similarity gives 7.70~\AA, and best-of-8 stem-selected gives 4.05~\AA, which is worse than the mean regression. The oracle loop has a median rank of 38 in stem-similarity order and enters the top 8 only 17\% of the time. The loop body is partly retrievable (full-loop oracle 1.71~\AA), but the apex is effectively unique in the framework frame, where no selector can place it.

\paragraph{The interface rides the same wall.} The interface metrics, such as Fnat, epitope F1, iRMSD, and DockQ, are all computed from the C$\alpha$--C$\alpha$ contact graph between predicted CDR positions and fixed antigen positions, so they are functions of the same coordinates. At that cutoff, a 2.4~\AA{} apex error flips borderline contacts. Sweeping the contact-recovery margin confirms the coupling: tightening the margin trades fnat up (0.565 to 0.612) against iRMSD down (1.537 to 1.710~\AA) along a DockQ contour near 0.67.

\paragraph{Implications.} These diagnostics make it safe to conclude that the position, framework flanks, antigen microenvironment, torsion co-design, and library retrieval are each dead ends for solving the apex region on this task. The sequence recovery on CDR-H3 is bounded by data rather than by architecture under these inputs, which is why \contact{} reports its gains on the structural and binding axes where the placement objective does move the frontier. The future work on raising sequence recovery further requires a fundamentally more accurate apex backbone generator or an external evolutionary signal such as a protein language model.

\end{document}